\documentclass[conference]{IEEEtran}
\IEEEoverridecommandlockouts
	
\usepackage{amsmath}
\usepackage{algorithmic}
\usepackage{graphicx}
\usepackage{textcomp}
\def\BibTeX{{\rm B\kern-.05em{\sc i\kern-.025em b}\kern-.08em
    T\kern-.1667em\lower.7ex\hbox{E}\kern-.125emX}}
\usepackage{siunitx}
\usepackage{amsfonts}
\usepackage[nocompress]{cite}
\usepackage{booktabs,multirow}
\usepackage{color}
\usepackage{tabularx}
\usepackage{orcidlink}
\usepackage{hyperref}
\usepackage{amssymb}
\hypersetup{
	colorlinks=true,
	linkcolor=blue, 
	citecolor=green,  
	urlcolor=blue   
}

\title{GVTNet: Graph Vision Transformer For Face Super-Resolution}

\author{	\IEEEauthorblockN{Chao Yang\orcidlink{0009-0009-1378-0644},
 	Yong Fan$^*$,
 	Cheng Lu\orcidlink{0009-0006-5324-5361}
 	,
 	Minghao Yuan,
 	Zhijing Yang\orcidlink{0009-0000-9865-0666}}
\IEEEauthorblockA{Southwest University of Science and Technology, China}}

\begin{document}
	
\maketitle

\begin{abstract} 
Recent advances in face super-resolution  research have utilized the Transformer architecture. This method processes the input image into a series of small patches. However, because of the strong correlation between different facial components in facial images. When it comes to super-resolution of low-resolution images, existing algorithms cannot handle the relationships between patches well, resulting in distorted facial components in the super-resolution results. To solve the problem, we propose a transformer architecture based on graph neural networks called graph vision transformer network. We treat each patch as a graph node and establish an adjacency matrix based on the information between patches. In this way, the patch only interacts between neighboring patches, further processing the relationship of facial components. Quantitative and visualization experiments have underscored the superiority of our algorithm over state-of-the-art techniques. Through detailed comparisons, we have demonstrated that our algorithm possesses more advanced super-resolution capabilities, particularly in enhancing facial components. The PyTorch code is available at \href{https://github.com/continueyang/GVTNet}{https://github.com/continueyang/GVTNet}
\end{abstract}
	\section{Introduction}
	\label{sec:intro}
	
	FSR (Face super-resolution) aims to improve the LR (low-resolution) to HR (high-resolution) face images, improving their clarity and detail levels\cite{survey}.  However, because of the irreversibility of the degradation process and the complexity and unknown properties of the degradation kernels in real-world scenarios. SR tasks are usually highly pathological, and an LR image can correspond to many HR images. These issues make it a long-standing and challenging research field in low-level visual representation. In the past, FSR methods based on prior guided methods\cite{fsrnet,urdgn}, attribute constraints\cite{att1}, and pure CNN (convolutional neural networks) have been proposed.\cite{rcan,aacnn,edsr}.

	Recently, the ViT (Vision Transformer) architecture has excelled in SR tasks (super-resolution), sparking significant interest and research in this field\cite{swinir,hat,DATdual,srformer,ctcnet}. It is well known that when an image is input into a ViT model, the ViT will decompose the image into small patches for processing \cite{vit,swin}, e.g., an image with a size of 128x128, into 8x8 patches. These patches are then flattened into 1D vectors as inputs for a multi-head self-attention mechanism. In previously developed ViT networks, a target with the same semantic information is divided into multiple patches to interact with other patches of the target that contain different semantics. However, because of the strong correlation between different facial components in facial images. This processing strategy is inflexible in mapping LR face images to HR face images.
	\begin{figure}[htb]
		\centering
		\centerline{\includegraphics[width=9cm,height=2.5cm]{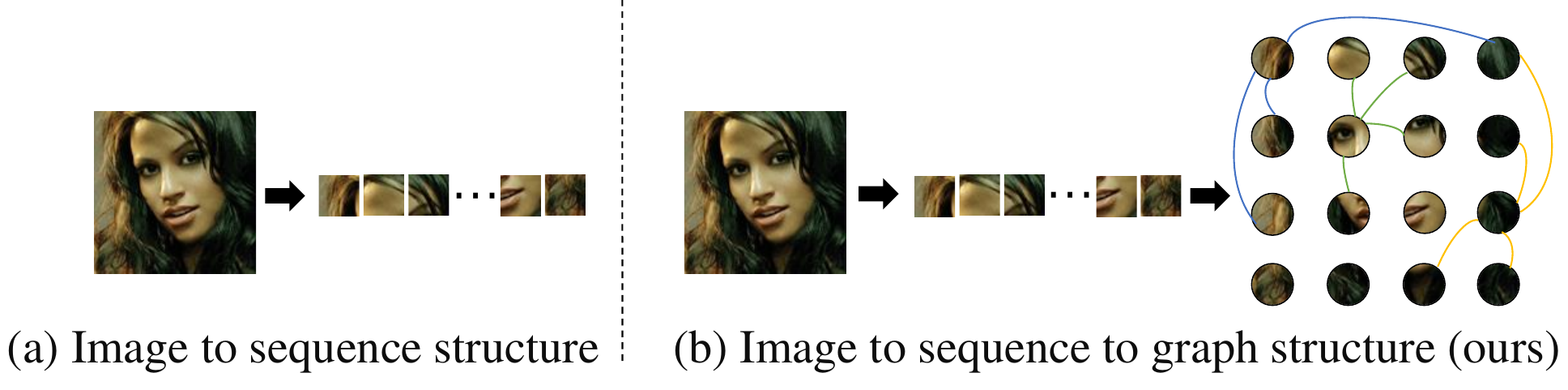}}
		\caption{The left image shows VIT's sequential structure, converting a 2D image into patches. On the right, our proposed structure links patches as graph nodes.}
		\label{fig:1}
	\end{figure}
Our visualization experiments on the results of SR reveal that existing methods often struggle to restore facial organs adequately when the resolution of the LR images is low or the SR multiplier is high. In some cases, the facial features in the output image are even distorted. We contend that there is a necessity to develop a model capable of modeling the connections between facial components more effectively.

	Inspired by GNN (graph neural networks)\cite{vsgnn,graphbert,transgnn}, Inspired by recent advancements in graph neural networks (GNNs) \cite{vsgnn,graphbert,transgnn}, we discovered that GNNs are capable of capturing intricate relationships within structured graph data more effectively. This capability endows them with a certain degree of "logicality" in terms of relational inference and structural perception. Therefore, we propose a transformer architecture using the idea of graph neural networks. And to our knowledge, our model is novel in the field of FSR. Specifically, in the ViT model, an image is partitioned into a series of patches and these patches can be considered nodes in the graph.  Building on this idea, we extend the attention mechanism to include graph structures. We use the information association between patches to create an adjacency matrix that captures complex spatial relationships, as shown in {\bf Fig.}\ref{fig:1}. Specifically, our model calculates the Minkowski distances between patches to construct an adjacency matrix that details the interrelationships between these patches. Using this adjacency matrix and graph feature aggregation, each node efficiently aggregates the features of its neighboring nodes to produce comprehensive neighborhood information during self-attention computation , which are crucial to efficiently processing the information of each sequence that contains different semantic features and the details of the patched faces.
	
	\begin{figure*}[htb]
		\centering
		\centerline{\includegraphics[width=18cm,height=4.5cm]{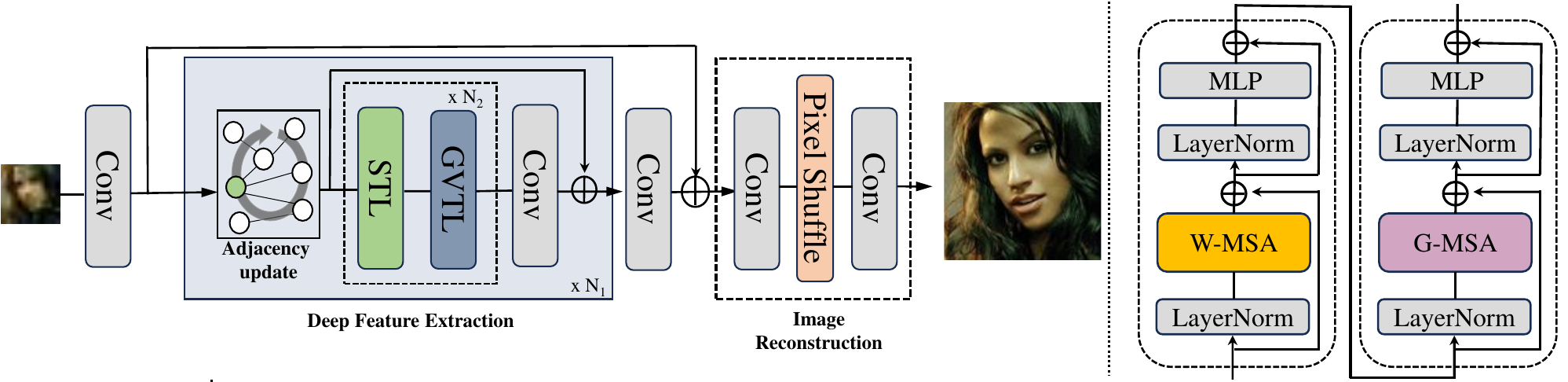}}
		\caption{The overall structure of our proposed algorithm. The left side is the overall framework of GVTNet, and the right side is the internal structure of DMB module.}
		\label{fig:2}
	\end{figure*}
	At the same time, although the use of our proposed GVT blocks can improve the performance of an FSR algorithm. But a network composed of pure GVT blocks inevitably misses part of the modeled information due to its neighbor selection mechanism. To balance this loss, we propose a dual-aggregate architecture with Swin and GVT blocks, which is modeled in two ways through the alternating use of traditional Swin layers and GVT layers. By this architecture, this model achieves a robust representation. Through quantitative, ablation, and visualization experiments, we show that our GVTNet achieves FSR results superior to those produced by current state-of-the-art (SOTA) methods and through detailed comparisons, we have shown that our algorithm has stronger SR capabilities for facial components.
	
	Overall, our contributions are threefold:
	
	i) We propose an FSR method using patch in VIT as a graph node. By aggregating neighbor node messages, we enhance the processing ability of the model for face images with strong correlation between different facial components.
    
	ii) We propose an FSR model, GVTNet, which utilizes dual aggregation mechanisms, Swin and GVT blocks, to produce powerful feature representations.
    
	iii) Our experimental results show that GVTNet outperforms current SOTA SR methods while maintaining low complexity and a small model size.
	
	\section{PROPOSED METHODOLOGY}
	\label{sec:pagestyle}
	GVTNet comprises two main components: deep feature extraction and image reconstruction; see {\bf Fig.} \ref{fig:2} for details. For LR facial images, the model initially generates shallow features via a simple 3x3 convolution. The output of the convolutional layer is then directed to a deep feature extraction module composed of $N_{1}$ GVT groups and adjacency update modules, incorporating residual links between these groups. Each group contains $N_{2}$ dual modeling blocks (DMB).
	
	After deep feature extraction is performed, the final output, $F_{D} \in \mathbb{R}^{H \times W \times C}$, is converted into an HR image through the reconstruction layer. This module employs the shuffling method for upsampling and convolution to aggregate features.
	\subsection{Adjacency Update Module}
	In our approach, ViT patches are treated as unique graph nodes with an adjacency matrix representing their neighborhood relationships. Our theoretical basis comes from the following work.
	
	1) A graph is a generalized data structure, and each patch in a sequence can be seen as a special case of the graph\cite{vsgnn}.
	
	2) GNNs and transformers share similarities. In natural language processing , sentences are often seen as fully connected graphs with each word linked to every other word. GNNs build features for each node (word) within such graphs to address NLP tasks. Unlike standard GNNs, which aggregate features from immediate neighbors, transformers consider an entire sentence as a neighborhood, aggregating features from all words in each layer\cite{graphbert}.
	
	3) A graph provides a unified representation for many interrelated data in the real world. It can model the different attribute information possessed by node entities, and each real-world object can be seen as an integration of different parts, such as the hair and facial features in a person's face image\cite{vsgnn,tsgnn2,tsgnn3}.
    
	This adjacency update module is placed in the deep feature extraction module after the shallow feature extraction module. After entering the deep feature extraction module, our adjacency matrix calculation module begins to collect neighbor information. During each iteration of the deep feature extraction module, the adjacency matrix calculation module recalculates and updates new neighbor information. The specific calculation process is as follows: 
    \begin{figure*}[htb]
		\centering
		\centerline{\includegraphics[width=17cm,height=5.5cm]{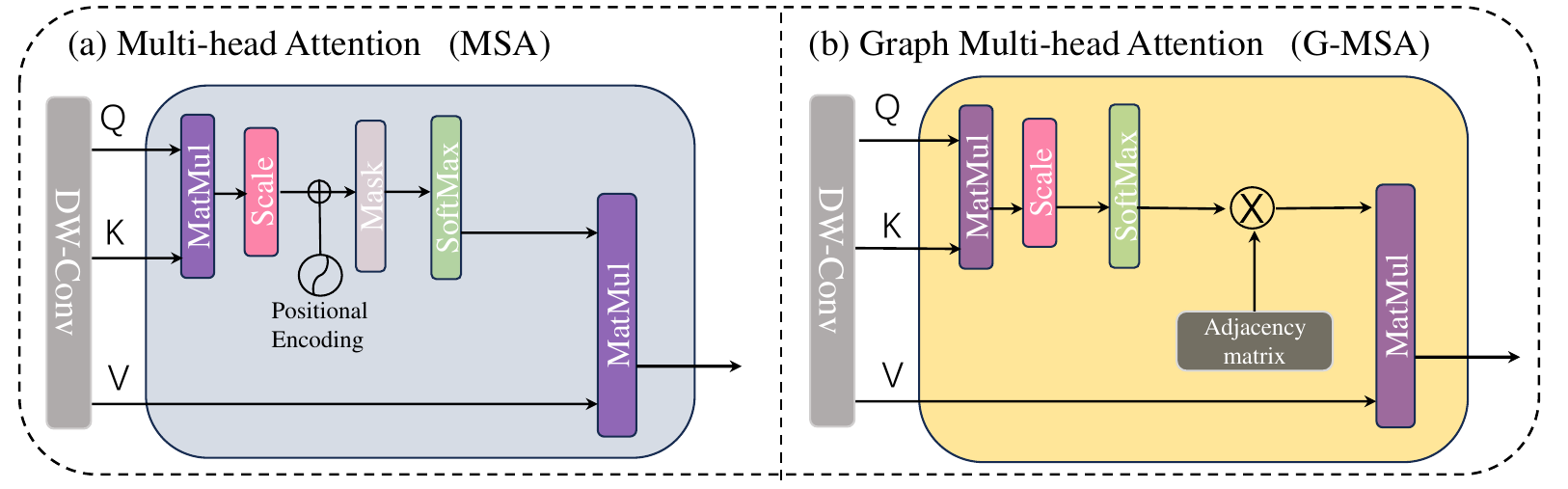}}
		\caption{The attention mechanism of our proposed GVTNet is compared with the internal structure of the traditional attention mechanism in SwinIR\cite{swinir}. The left side is the traditional attention, and the right side is our proposed G-WSA.}
		\label{fig:3}
	\end{figure*}
    
	For the set of input patch vectors $Z = \{z_1, z_2, \ldots, z_n\}$, we compute the Minkowski distance between each pair of vectors. The formula for doing so is as follows:
    \begin{equation}
    \begin{gathered}
            A = \begin{cases} 1, & \text{if } D_{min}(z_i, z_j) > T \\ 0, & \text{otherwise} \end{cases}
            \end{gathered}
	\end{equation}
    
    A is the adjacency matrix that we use to record information about neighbors and T is the threshold used to select neighbors; nodes with mutual distance greater than T are recorded as neighbors.
    
    \begin{equation}
    \begin{gathered}
			D_{min}(\mathbf{z_i}, \mathbf{z_j}) = \left( \sum_{i=1}^{n} |z_i - z_j|^p \right)^{\frac{1}{p}}
    \end{gathered}
    \end{equation}
			
	where $i,j = 1,2,\ldots,n$ and $i \neq j$.
	
	Based on the distances between the patches, adding an edge $\mathbf e_{ij}$ from $\mathbf{z_i}$ to $\mathbf{z_j}$ within the adjacency matrix $A$. Upon receiving the output from the preceding layer, the GVT group of each layer executes the module to refresh the neighbor information.

	
	\subsection{Dual Modeling Blocks (DMB)}
	Networks composed solely of GVT layers may miss certain information due to their neighbor selection mechanisms. To mitigate this issue, we introduce a dual modeling architecture combining Swin and GVT layers. This strategy employs both layers alternately for bidirectional mapping, enhancing the representational capabilities of the SR model. Specifically, the overall architecture of each of the two modules is composed of a MLP (multilayer perceptron), one attention layer, and two layer normalization layers. The main difference lies within the attention calculation process of the GVT layer.
	
	\vspace{10pt}
	\noindent {\bf Graph Vision Transfromer Layer}: In this layer, we use the previously calculated adjacency matrix in calculating attention; see {\bf Fig.} \ref{fig:3} for details of the specific structure. For a given $X \in \mathbb{R}^{H \times W \times C_{i n}}$, (H, W and $C_{i n}$ are the height, width, and number of input channels, respectively), we first use a depthwise separable convolution \cite{dwconv} to generate the query, key and value matrices (denoted Q, K and V, respectively). Then, after calculating the attention score matrix for each patch for the other patches , we use our adjacency matrix to select the attention score for each patch. Because both the attention matrix and the adjacency matrix are obtained through matrix multiplication. Therefore, neighbor information can be applied to attention operations by multiplication of dots. After this operation, our attention information is only transmitted between neighboring nodes. The whole calculation process is as follows.
	\begin{equation}
		Q=X P_Q, \quad K=X P_K, \quad V=X P_V,
	\end{equation}
	where $P_Q$, $P_K$ and $P_V$ are projection matrices shared between different windows. The method we use is depthwise separable convolution.
	
	\begin{equation}
		\begin{gathered}
			X=\operatorname{SoftMax}\left(Q K^T \cdot A/ \sqrt{d}\right) V \\
		\end{gathered}
	\end{equation}
	where $A$ represents the adjacency matrix that we calculate before the module starts, and we aggregate the neighbour node information via point multiplication.
	
	\noindent {\bf Swin Tranformer layer} : Before each GVT layer, we retain a classic STL , and the overall structure is derived from SwinIR\cite{swinir}; see {\bf Fig.}\ref{fig:3} for details of the specific structure.  The whole calculation process is as follows:
	\begin{equation}
		\begin{gathered}
			X=\operatorname{SoftMax}\left(Q K^T / \sqrt{d}+B+Mask\right) V \\
		\end{gathered}
	\end{equation}
	where B is the encoding of the relative position that can be learned. The MASK in question refers to the window displacement mechanism in the SwinIR\cite{swin}. In order for it not to interfere with our application of the adjacency matrix, we turn it off in the G-MSA module.

	\section{Experiments}
	\subsection{Experimental Settings}
Our experiment used the CelebA\cite{celeba} and Helen\cite{helen} datasets, resizing face images to $128 \times 128$ pixels. For CelebA, we used 168,854 samples for training, 100 for validation, and 1,000 for testing. For GVT groups, the numbers are configured to 6. The window size and number of attention heads are set to 8 and 6. We implemented the model using the PyTorch framework. Our model optimization uses the Adam optimizer, with$ \beta_1$ set to 0.9 and $\beta_2$ set to 0.99. We set the learning rate at $2 \times 10^{-4}$. Our experiments were conducted with an NVIDIA RTX A6000 graphics card. To evaluate the quality of SR results, we employed objective image quality indicators: peak signal-to-noise ratio (PSNR) \cite{psnr} and structural similarity (SSIM) \cite{ssim}. More experimental details can be found in our open-source code.

	\subsection{Ablation Study}
	
	\begin{table}[h]
		\centering
				\caption{Ablation experiments on the effectiveness of our proposed algorithm and the influence of hyperparameters on experimental results}
		          \label{label:1}
		\vspace{\baselineskip}
		\resizebox{0.35\textwidth}{!}{
			\begin{tabular}{c c |c |c c}
				
				\hline
				Threshold  & p & Methods & PSNR  & SSIM  \\ \hline 
				-            & -       & Baseline& 27.83             & 0.8133           \\ 
				0.85             & 2      & w/o STL& 27.93             & 0.8158           \\ 
				0.85             & 2      & DMB& 27.95             & 0.8161           \\ 
				0.85             & 2      & w/o DW-conv & 27.92             & 0.8150           \\
				0.85             & 1      &- & 27.91             & 0.8151           \\ 
				0.85              & $\infty$    & - & 27.93             & 0.8156\\
				0.75              & 2     & - & 27.95             & 0.8163\\
				0.60              & 2      & - & 27.94             & 0.8155\\ \hline
			\end{tabular}
		}
        
	\end{table}
	To verify the efficacy of our GVT module and DMB, we performed an ablation study, as detailed in the second row of {\bf Table }\ref{label:1}. Substituting all STLs with GVTLs in the baseline model resulted in a 0.12 dB improvement, demonstrating that GVTNet provided a significant SR enhancement. However, to supplement the departmental information missing from the GVT layer. In order to verify that our DMB can solve this problem, we performed ablation experiments, as shown in the third line of {\bf Table }\ref{label:1}. After adding this module, the performance was upgraded to 0.14 db. In addition, we conducted experiments to determine how our model performed without depthwise separable convolution, as shown in the fourth row of {\bf Table }\ref{label:1}.
	\begin{table*}[htbp]
		\centering
		\caption{Quantitative comparison with state-of-the-art methods on benchmark
			datasets. The top two results are marked in \textcolor{red}{red} and \textcolor{blue}{blue}}
		\label{table:2}
		\sisetup{table-format=2.2} 
		\begin{tabular}{
				l
				S[table-format=2.2]
				S[table-format=1.4]
				S[table-format=2.2]
				S[table-format=1.4]
				S[table-format=2.2]
				S[table-format=1.4]
				S[table-format=2.2]
				S[table-format=1.4]
				c
				c
			}
			\toprule
			{Methods} & \multicolumn{2}{c}{CelebA\cite{celeba}×4} & \multicolumn{2}{c}{CelebA\cite{celeba}×8} &  \multicolumn{2}{c}{Helen\cite{helen}×4} & \multicolumn{2}{c}{Helen\cite{helen}×8}& {Params}  \\
			\cmidrule(lr){2-3} \cmidrule(lr){4-5} \cmidrule(lr){6-7} \cmidrule(lr){8-9}
			& {PSNR↑} & {SSIM↑} & {PSNR↑} & {SSIM↑}  &  {PSNR↑} & {SSIM↑} & {PSNR↑} & {SSIM↑} \\

			\midrule
			\multicolumn{9}{c}{Prior-guided FSR Methods} \\
			\hline
			DIC \cite{DIC} & 31.44 & 0.9091 & 27.41 & 0.8023  &31.62&0.9127&27.54&0.8227& 20.8M \\
			FSRNet\cite{fsrnet}&31.46& 0.9083 & 26.66& 0.7718 &31.59&0.9122&25.45&0.7364&3.1M \\
			\midrule
			\multicolumn{9}{c}{General Image Super-Resolution Methods} \\
			\hline
			EDSR\cite{edsr}&31.57& 0.9011 & 27.24& 0.7900 &32.33&0.9227&27.49&0.8184&17.53M\\
			RCAN\cite{rcan}&31.77& 0.9020 & 27.30& 0.7824 &32.42&0.9236&27.50&0.8231&15.00M\\
			SwinIR\cite{swinir}&31.32& 0.9097 & 27.83& 0.8133 &32.81&0.9294&27.42&0.8170&12.05M\\
			NLSN\cite{nlsn}&32.08& 0.9090 & 27.45& 0.8043 &32.24&0.9217&27.57&0.8210&43.40M\\
			\midrule
			\multicolumn{9}{c}{General FSR Methods} \\
			\hline
			SPARNet \cite{sparnet} & 31.71 & 0.9021 & 27.44 & 0.8047 &32.37&0.9235&27.73&0.8227 & 10.0M \\
            SFMNet\cite{sfmnet}&32.01& \textcolor{red}{0.9175} & 27.56& 0.8074 &32.51&0.9187&27.22&0.8141&8.1M\\
            W-Net\cite{wu}&31.77& 0.9032 & 27.54& 0.8041 &32.32&0.9187&27.26&0.8121&--\\
			GVTNet(Ours)&\textcolor{blue}{32.52}& 0.9144 & \textcolor{blue}{27.95}&\textcolor{blue}{ 0.8061} &\textcolor{blue}{33.13}&\textcolor{blue}{0.9333}&\textcolor{blue}{27.77}&\textcolor{blue}{0.8296}&12.17M\\
			GVTNet+(Ours)&\textcolor{red}{32.62}& \textcolor{blue}{0.9155} & \textcolor{red}{28.06}& \textcolor{red}{0.8189} &\textcolor{red}{33.28}&\textcolor{red}{0.9347}&\textcolor{red}{28.00}&\textcolor{red}{0.8351}&12.17M\\
			\bottomrule
			
		\end{tabular}
		
	\end{table*}
	
	
	\begin{figure}[htb]
		\centering
		\centerline{\includegraphics[width=9cm,height=4.5cm]{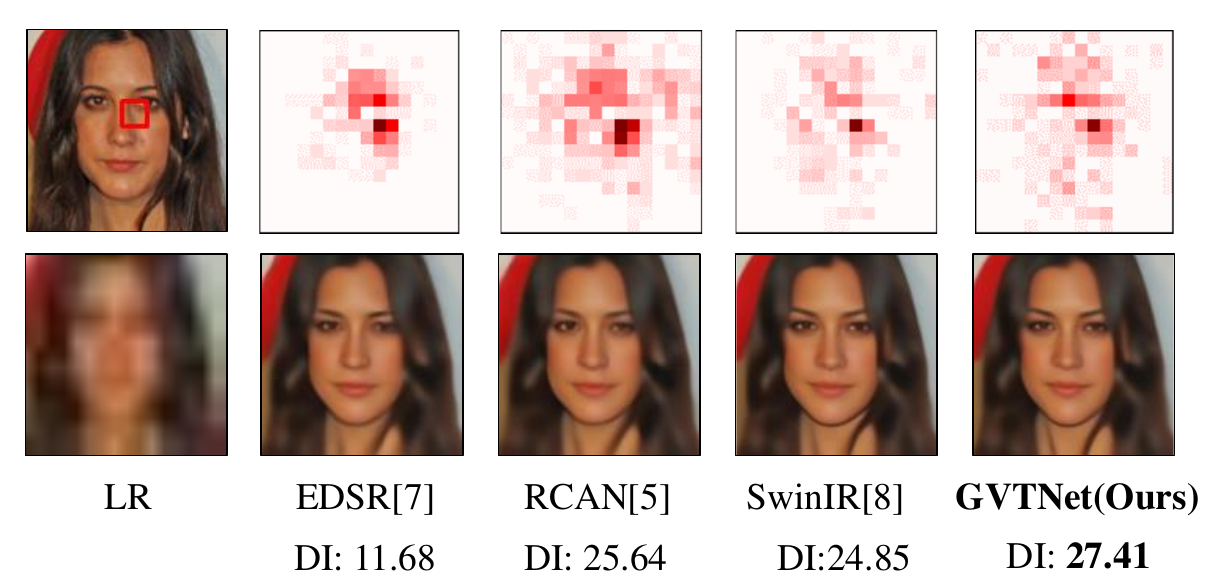}}
		\caption{The visual analysis results of LAM\cite{LAM} , The DI value represents the magnitude of the range of information used by the model to recover the target.}
		\label{fig:4}

	\end{figure}

	\noindent {\bf Hyperparameter effects}: Upon entering the model, the input image will undergo a convolutional process, resulting in a mapping of the number of channels from three to hundreds of dimensions within a high-dimensional space. The optimal distance measurement method between the image patches in this space to measure the relationship of information between nodes remains unknown. To verify the effect of using different p-values for the Minkowski distance to select neighboring nodes and improve the SR effect, we performed ablation experiments on the three common parameters $p = 1, 2,\infty$ in {\bf Table }\textcolor{blue}{I}. The experimental results showed that different neighbor selection schemes had little effect on the experimental results and $p = 2$ was the best. At this point, the Minkowski distance is equal to the Euclidean distance.

	The threshold is a major hyperparameter in our proposed algorithm. The larger the threshold, the fewer neighbor nodes the model selects. The smaller the threshold, the more neighbor nodes the model selects. We designed experiments to verify the impacts of different thresholds on the results. In {\bf Table }\ref{label:1}, the experimental results show that the use of too many or too few neighboring nodes produced poor experimental results. Finally, we chose 0.75 as our threshold.
	\begin{figure}[htb]
		\centering
	   
		\centerline{\includegraphics[width=8.5cm,height=4.5cm]{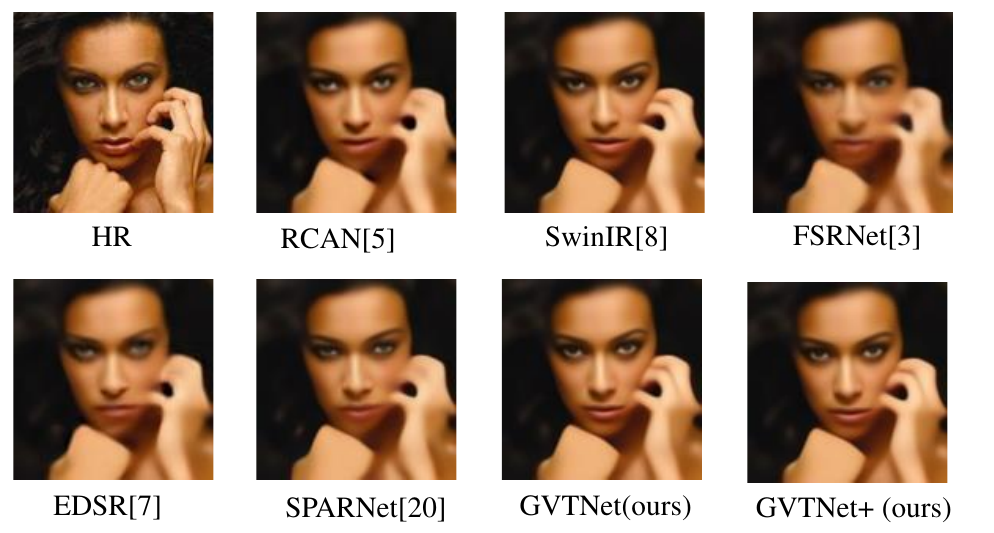}}
	\caption{Test results visualized at X8 super-resolution on the Celeba\cite{celeba} test set.}
	  \label{fig:5}
	\end{figure}

	\subsection{Comparisons With State-of-the-Art Methods}
    Currently, mainstream FSR algorithms can be categorized into three types: General FSR Methods, Prior-Guided FSR Methods, and General Image SR Methods.
General FSR Methods: These approaches focus on identifying and refining specific facial details to produce recognizable high-quality images.
Prior-Guided FSR Methods: Building on general FSR techniques, these methods incorporate additional facial information, such as landmarks, to achieve more precise and realistic reconstructions. This structured knowledge helps the model better predict and recreate facial features.
General Image SR Methods: Designed to enhance the resolution of a wide range of image types, not just faces, these techniques utilize CNN (Convolutional Neural Networks) and GAN (Generative Adversarial Networks) to improve details across diverse content.

	We compared the GVTNet model with the three SOTA SR methods mentioned above, including (SPARNet\cite{sparnet}, DIC\cite{DIC} , FSRNet\cite{fsrnet},EDSR\cite{edsr}, RCAN\cite{rcan}, SwinIR\cite{swinir}, NLSN\cite{nlsn} and SFMNet\cite{sfmnet}). Consistent with previous studies \cite{swinir}, we adopted a self-integration strategy during the test, which is represented by the symbol ' + '. {\bf Table.} \textcolor{blue}{II} provides a quantitative comparison, showing the SR results of the image obtained with factors of × 4 and × 8. Our GVT was superior to the comparison methods in terms of overall performance. At the same time, the performance of GVTNet + was better than that of the previous methods. Even without the self-integration strategy, compared to the baseline SwinIR model, our GVTNet achieved a significant gain (×4) on the HELEN dataset, with an improvement of 0.32 dB. For the SSIM metric, GVTNet achieved the best performance except celeba (x4), and our model improved by around 0.01 in other datasets. These quantitative results show that the graph node neighbor selection strategy implemented through GVTNet can effectively provide improved image SR quality.
	\begin{figure}[htb]
		\centering
		\centerline{\includegraphics[width=8.5cm,height=7cm]{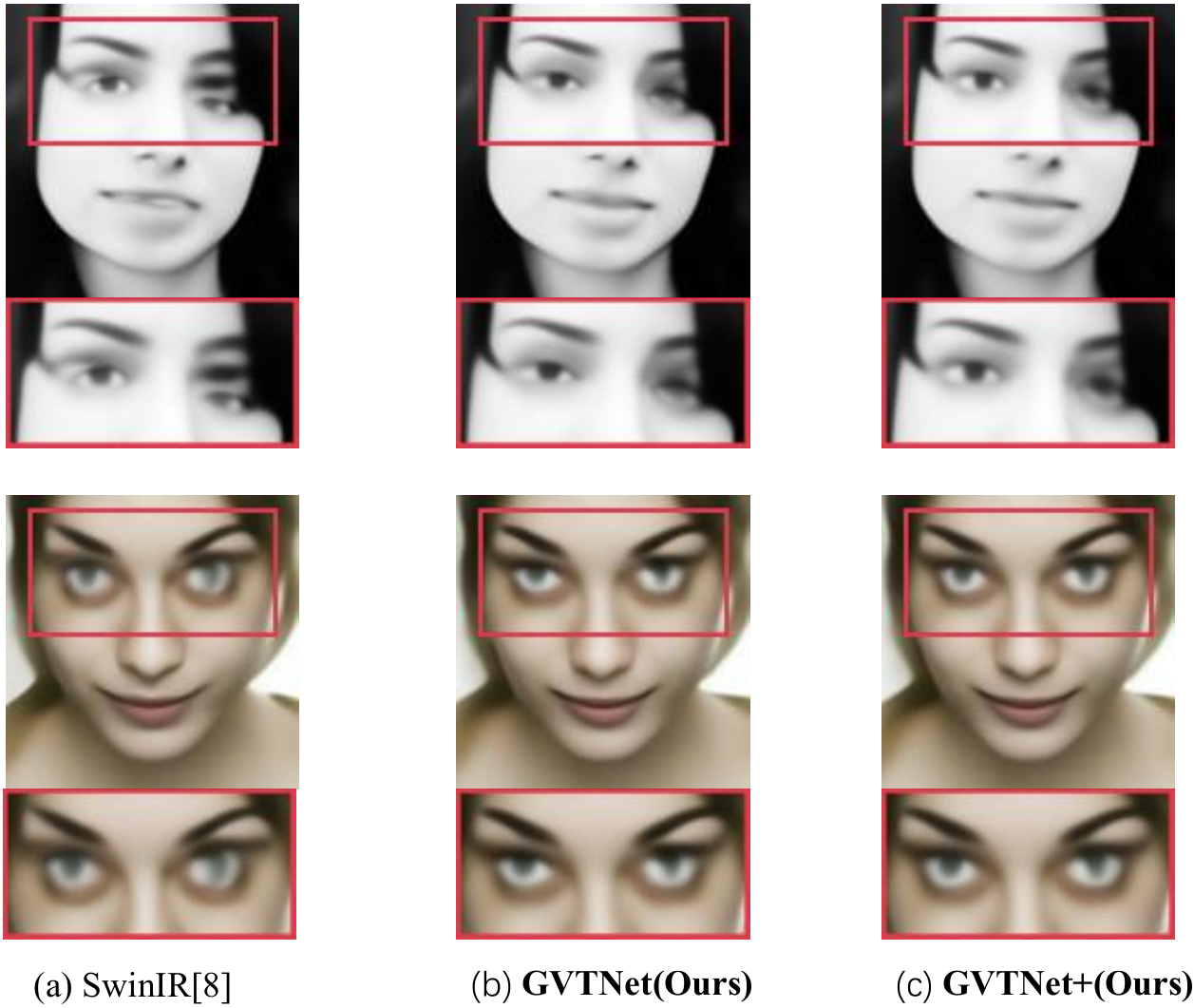}}
		\caption{Visual comparison experimental results of detail recovery ability with baseline model.}
		\label{fig:6}
	\end{figure}
    
	\noindent {\bf Visualization experiment}: To further verify the ability of our algorithm to generate faces, we provide a visual comparison with the current SOTA algorithm in {\bf Fig.} \ref{fig:5}. Visualization experiments revealed that our algorithm accurately restored both the facial structure of the target character and the image details. In particular, our algorithm demonstrated substantial improvements, such as enhanced hand details, as shown in {\bf Fig.}\ref{fig:5}. Since the proposed model can interact with information from different parts of the face , it reduces the incompleteness of recovered faces due to the severe lack of information in LR images. 
    
    The same as the visualization experiment of previous work\cite{gnn2024,gnn3}, in order to demonstrate the recovery capacity of detail of our model, we compared the recovery results with our baseline model by visualizing the details in {\bf Fig.}\ref{fig:6}. From the figure, it can be seen that our model reduces the facial errors in recovery. This shows that our model has a better understanding of the semantic relevance of each part of the face.
    
	Furthermore, we used the LAM\cite{LAM} attribution analysis method to evaluate the efficiency of pixel utilization of our algorithm for facial image restoration, as illustrated in {\bf Fig.}\ref{fig:4}. LAM is a commonly used interpretable visualization algorithm for SR, which allows one to see which information from other regions the algorithm used when restoring certain areas. In previous work, such as HAT\cite{hat,DIC}, a statement was proposed that the wider the surrounding pixels used to restore the target, the stronger the performance and generalization of the algorithm. The experimental results indicate that our algorithm effectively captured the semantic information of the target image. At the same time, the experimental results also show that our algorithm has the largest pixel utilization range. This shows that our algorithm has a good understanding of the semantic information contained in different parts of the face.
	\section{Conclusion}
	Inspired by the idea of a GNN, this paper proposes a new perspective for processing images in a transformer-based FSR algorithm. A ViT treats the image to be processed as a patch sequence. We regard each patch as a graph node, use the Minkowski distance to filter neighboring nodes for each graph node, and create an adjacency matrix for modeling the mapping process. At the same time, to balance the inevitable loss of information due to the neighbor selection mechanism used and to enhance the diversity of the output information, we propose a dual modeling module. Many quantitative and visualization experiments comparing the proposed approach with the SOTA algorithms prove that our algorithm restores more detailed features while maintaining a smaller model size.

\vfill\clearpage

\bibliographystyle{IEEEbib}
\bibliography{IEEEabrv}
	
\end{document}